\documentclass[runningheads]{llncs}
\usepackage{graphicx,multirow}
\begin{document}
\title{Compressive sensing based privacy for fall detection}
%
%

\author{
Ronak Gupta\inst{1} \and
Prashant Anand\inst{1} \and
Santanu Chaudhury\inst{1,2}
Brejesh Lall\inst{1} \and
Sanjay Singh \inst{3}
}


%
\authorrunning{Ronak et al.}
%

\institute{
Department of Electrical Engineering, Indian Institute of Technology Delhi 
\email{ronakgupta143@gmail.com}\\ \and
Indian Institute of Technology Jodhpur \and
Cognitive Computing Group, CSIR-CEERI, Pilani, India\\
}
\maketitle              
\begin{abstract}
Fall detection holds immense importance in the field of healthcare, where timely detection allows for instant medical assistance. In this context, we propose a 3D ConvNet architecture which consists of 3D Inception modules for fall detection. The proposed architecture is a custom version of Inflated 3D (I3D) architecture, that takes compressed measurements of video sequence as spatio-temporal input, obtained from compressive sensing framework, rather than video sequence as input, as in the case of I3D convolutional neural network. This is adopted since privacy raises a huge concern for patients being monitored through these RGB cameras. The proposed framework for fall detection is flexible enough with respect to a wide variety of measurement matrices. Ten action classes randomly selected from Kinetics-400 with no fall examples, are employed to train our 3D ConvNet post compressive sensing with different types of sensing matrices on the original video clips. Our results show that 3D ConvNet performance remains unchanged with different sensing matrices. Also, the performance obtained with Kinetics pre-trained 3D ConvNet on compressively sensed fall videos from benchmark datasets is better than the state-of-the-art techniques.
\end{abstract}
\section{Introduction}

As per WHO report \cite{krishnaswamy2003falls}, India is the second most populous country in the world with more than 75 million people lying in the age group of more than 60 years. Human fall is a serious problem concerning people with this age group and is considered as one of the "Geriatric Giants" \cite{krishnaswamy2003falls}. Therefore, to address this issue, the need for intelligent monitoring system of the elderly people has risen over the past years. The precise objective for these systems is to automatically detect falls while minimizing false negatives and then to intimate the caregivers/family members.

Several deep learning based fall detection techniques \cite{lu2019deep,feng2018spatio,adhikari2017activity,nogas2018fall} have been presented and for generalization few depend on large action recognition datasets for pre-training. In ~\cite{lu2019deep} authors proposed a scheme for fall detection through ambient camera, where they employed 3D convolutional neural network (3D CNN) to obtain coarse spatio-temporal features, This was followed by Long short-term memory (LSTM) based visual attention mechanism to extract the motion information encoded within the region of interest from coarse spatio-temporal features of the video sequence. The kinetic database Sports-1M which does not have fall data was used for training the 3DCNN. In ~\cite{adhikari2017activity} fall events are detected as a series of sequential change in human pose and these different poses are recognized using CNN. They tried different input image combinations of RGB, Depth, background subtracted RGB to name a few as input to the CNN. Their focus was on human silhouette extracts for recognizing human pose for fall detection.

In this paper, we propose 3D ConvNet architecture which consists of 3D Inception modules for the task of fall detection. The architecture takes spatio-temporal input in compressed domain, rather than spatio-temporal input in image domain as done in Inflated 3D (I3D) architecture. The compressive sensing captures the measurements which are then used for performing classification as a fall or other daily activities (labelled as non fall). In visual systems, while training the fall data is usually generated by simulated falls under a variety of circumstances, that makes it difficult to obtain large quantity of training instances and thus trained classifier has high chance of overfitting the training data. Also, since both the fall dataset used for experiments do not have sufficient training samples, we pre-train the architecture on action recognition datasets for learning better representation of the input videos. This significantly improves the generalization of the deep neural network by giving good detection rates~\cite{tran2015learning,carreira2017quo}.

The authors adopt compressive sensing step in the recognition framework which render the compressive samples visually imperceptible. This is  essential in circumstances where one might prefer a system which doesn't disclose their identity and capturing all personal activities/details via  visual systems/cameras used for detecting falls poses a serious threat to one's privacy. Compressive sensing demonstrates that a signal that is K-sparse in one basis called sparsity basis can be recovered or classified from K linear projections onto a second basis. The latter is called measurement basis which is incoherent with the first. While the measurement process is linear, the reconstruction or classification process has to be done through non linear transformations. It is also a well known fact that the compressive samples of images/video frames containing personal information can essentially be used to achieve privacy. This is because CS transformation is viewed as a symmetric cipher resulting in computational secrecy when the secret sensing matrix is unknown to the adversary \cite{orsdemir2008security,rachlin2008secrecy,hu2017compressive}. 

Although, several privacy based intelligent systems for fall detection have been designed in the past \cite{mirmahboub2013automatic}. These systems employ action recognition algorithms which run directly on the camera monitoring the person thus enhancing privacy. Their deployment is done in such a manner that only the fall alarms are transmitted but the the video frames are not. Other popular systems \cite{nogas2018fall} are usually based on thermal heat- maps although capable of masking the person’s identity effectively but are an expensive option. The earlier in-house implementation will be problematic to update when new instances are available \cite{mirmahboub2013automatic}. In contrast to the aforementioned approaches, compressive sensing field suggests that a small group of linear projections of a compressible signal contains enough information for reconstruction, classification and processing \cite{kulkarni2016reconstruction,kulkarni2012recurrence,xu2018csvideonet,wakin2006architecture,gao2015block,dinh2013measurement,kulkarni2016reconnet}.


\section{Related works}
Existing non-deep learning fall detection techniques depends on extracting the person (foreground) first, which is highly influenced by image noise (background), illumination variation and occlusion. In ~\cite{rougier2011robust} authors presented the fall detection by quantifying human shape deformation. For human shape change analysis, they extract and compare two consecutive silhouettes of a person. The landmarks/edge points extracted from silhouette are then matched through video sequence to quantify the silhouette deformation. They compare the mean matching cost of silhouette landmarks and the full Procrustes distance ~\cite{dryden2014shape} as body shape deformation measures. Based on these shape deformation measures during the fall followed by a lack of significant movement after the fall are fed to Gaussian Mixture Model (GMM) to classify the different activities as fall or not. In ~\cite{mirmahboub2013automatic} the authors presented a fall detection system that uses silhouette area as a feature. Their approach works irrespective of the direction of the movement of the person with respect to the camera. They present a mathematical analysis to confirm the relation between silhouette area and a fall event. The classification is done separately based on the variations of silhouette area as features for SVM classifier.

In ~\cite{feng2018spatio} authors have proposed a spatial-temporal fall detection method, which can present specific spatial and temporal locations of fall events in complex scenes. In their method, an object detector YOLO v3 ~\cite{redmon2018yolov3} is used for person detection, later a deep learning based method for multi-object tracking is used. The features from the tracker are fed to an attention guided LSTM model to detect specific fall events. In ~\cite{nogas2018fall} the authors presented the use of thermal camera for fall detection which is privacy preserving as it effectively masks the identity of those being monitored. They formulated the fall detection problem as an anomaly detection problem and used Convolutional LSTM Autoencoders to identify unseen falls.

In compressive sensing, random Gaussian matrix or random Bernoulli matrix has been widely used to generate linear measurements of natural images, frames of video, etc. ~\cite{gao2015block}. In practice there are several problems with GRM such as GRM is non-sparse and complicated, and hence highly computational complex and highly difficult in hardware implementation. The other issue is that the measurements generated by GRM are random, neither are data-driven nor adjacent measurements have enough correlation. In literature other measurement matrices have been proposed to solve the above issues. In ~\cite{dinh2013measurement}, the authors proposed structural measurement matrix (SMM) to achieve a better Rate-Distortion performance in CS based image coding, in which the image is sampled by small blocks for better measurement coding while CS recovery can be performed in large blocks for better quality of recovered images. Their method of measurement coding with SMM, helps exploit the spatial correlation in measurement domain, which is represented by directional pixel behaviour (i.e object edges), that improves measurement prediction scheme and reconstructed with large blocks spliced from small correlated blocks improves CS recovery. In ~\cite{gao2015block}, the authors proposed a novel local structural measurement matrix (LSMM) for block-based CS coding of natural images by utilizing the local smooth property of images. Their proposed LSMM is a highly sparse matrix and the adjacent measurement elements generated by LSMM have high correlation that has been shown to improve the coding efficiency of spatial information.

Outline of the paper is as follows: Section \ref{sec:method} introduces methodology to solve the problem and the proposed architecture. Section \ref{sec:results} presents experimental results to show the effectiveness of the framework and Section ~\ref{sec:conclusion} concludes the paper.

\section{Methodology}
\label{sec:method}

\begin{figure}[htbp]
    \centering
    \includegraphics[width=12.5cm]{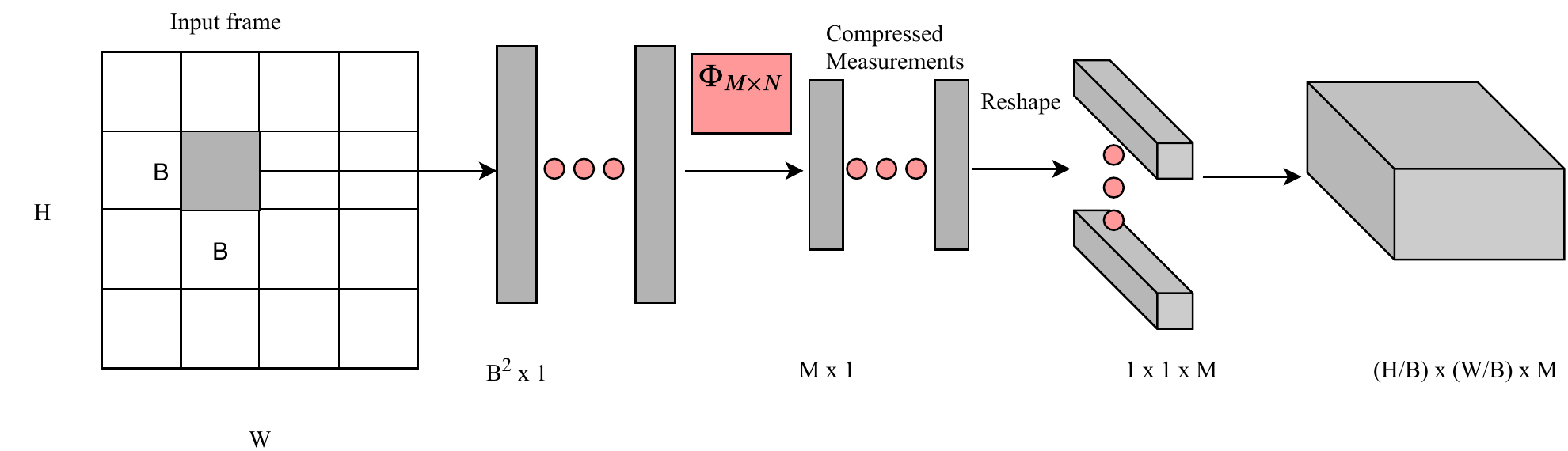}
    \caption{Compression technique}
    \label{fig:compression_tech}
\end{figure}

\begin{figure}[htbp]
    \centering
    \includegraphics[width=12.5cm]{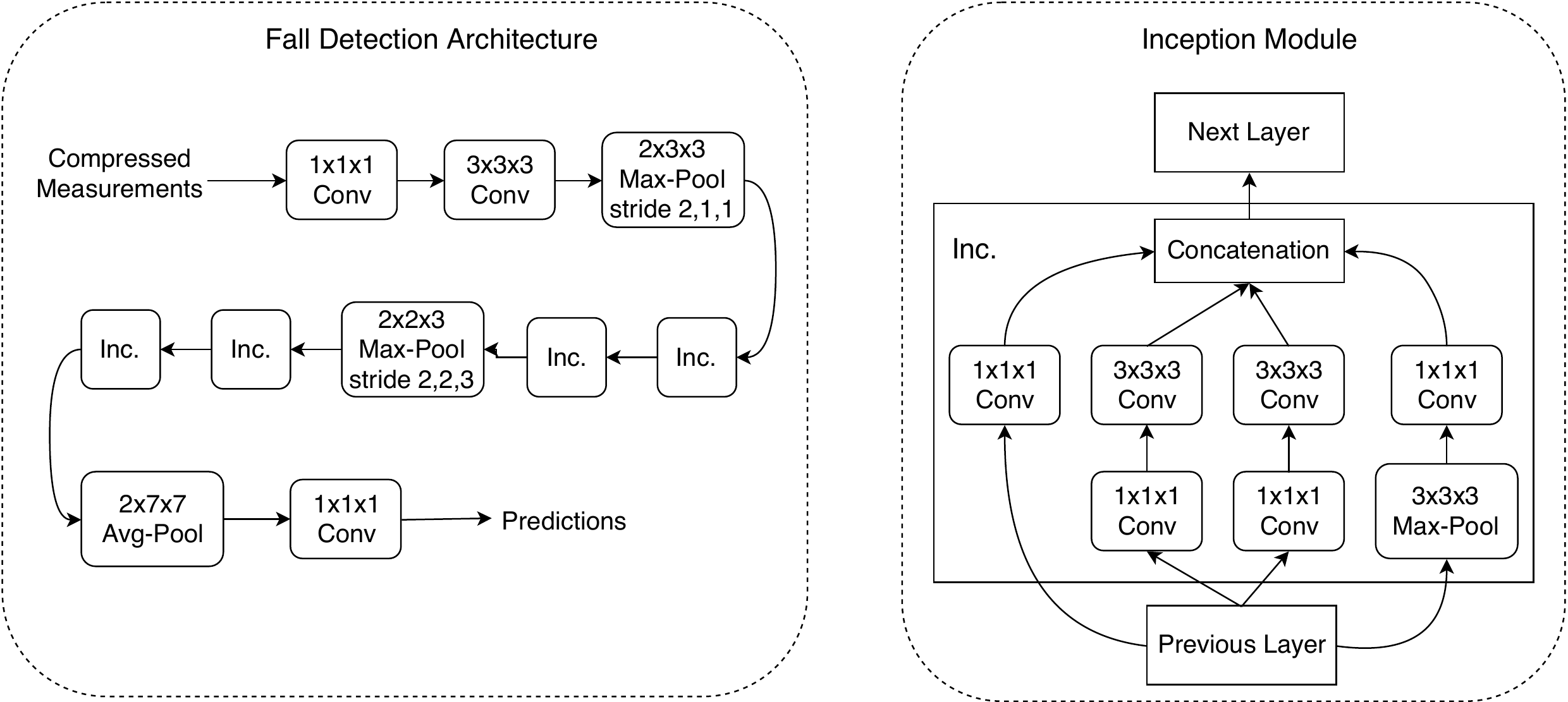}
    \caption{Fall detection architecture}
    \label{fig:fall_det_arch}
\end{figure}

We use 3D ConvNet which includes submodules designed from Inception-V1 network architecture for fall detection. The submodules present in Inception-V1 architecture are inflated as done in I3D Convolutional neural network ~\cite{carreira2017quo} to construct 3D ConvNet. The inflated Inception-V1 modules are found to be more effective in action recognition compared to VGG-style 3D CNN~\cite{carreira2017quo}. There are four inflated Inception submodules in our 3D ConvNet architecture. For fall detection, our 3D ConvNet takes compressed measurements of video sequence as spatio-temporal input, obtained from compressive sensing framework (as shown in Figure \ref{fig:compression_tech}), rather than video sequence as input, as in the case of I3D convolutional neural networks. Here, the compressed measurements for RGB frames of given video sequence are stacked together along the color (RGB) channel dimension. Figure \ref{fig:fall_det_arch} shows the fall detection architecture. 

We adopt a compressive sensing step in the recognition framework which render the compressive samples visually imperceptible, a necessity for privacy. When block based compressive sensing is performed over video frame, we get compressed measurements for the corresponding block. If the dimension of block is $N (=B^2)$ and when it is multiplied with a sensing matrix of size $M$x$N$, we get $M$ measurements and the compression ratio is defined as $r=\frac{N}{M}$. The compressed measurement vectors obtained for corresponding blocks in a frame, are arranged across channel dimension as shown in figure before given as input to fall detection architecture. Hence, when compressive sensing is applied to the frame at block level, the output compressed representation will have spatial dimension depending on the number of blocks in video frame and the channel dimension depending on the compression ratio. Similar rearrangement of images or video frame is also performed in the inverse pixel shuffling operation present in sub-convolutional layer of image or video super-resolution frameworks ~\cite{shi2016real}. The difference between their inverse pixel shuffling operation is that it does not involve dimensionality reduction. Moreover, the linear transformation involved in CS of the video frame blocks into compressed measurements makes rearrangement of the measurements back to the input frame difficult compared to pixel shuffling in sub-convolutional layer.

We show that our CS based privacy for fall detection architecture can work with different compressive sensing matrices. Random Gaussian matrix or random Bernoulli matrix has been used to generate random linear measurements of the video frame blocks. We have also used structural measurement matrix and local structural measurement matrix which exploits intra-block correlation in spatial domain.

\section{Experimental Results}
\label{sec:results}
In this section we report performance of our framework over action recognition and fall datasets with a wide variety of sensing matrices. Once our 3D ConvNet is trained on action recognition dataset, we fine-tune the network for fall detection dataset.
\subsection{Fall and Action Datasets}
\begin{figure}[]
    \centering
    \includegraphics[width=\textwidth]{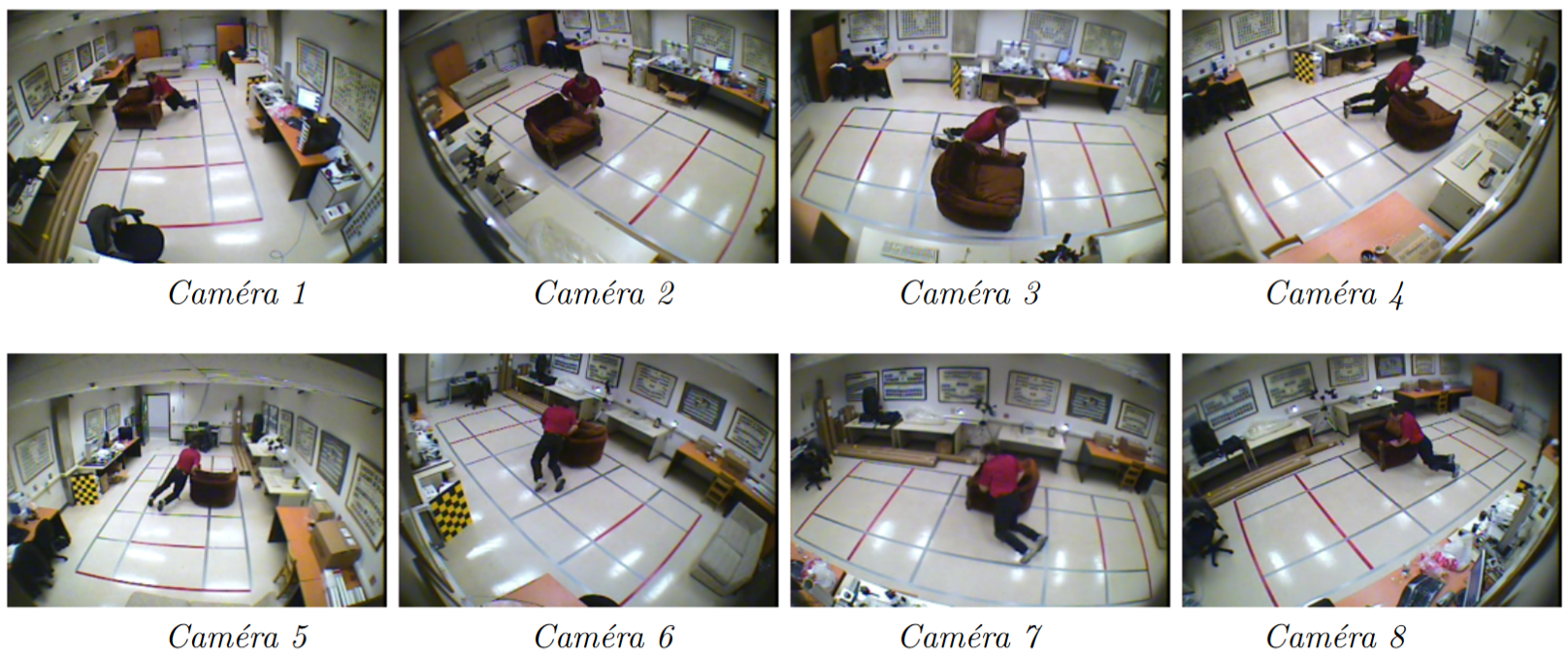}
    \caption{Fall example from multiple cameras ~\cite{auvinet2010multiple}}
    \label{fig:falldata}
\end{figure}

In ~\cite{auvinet2010multiple}, the authors collected a dataset of fall and normal activities from a calibrated Multi-camera system, of eight inexpensive IP cameras with a wide angle to cover the whole room. There are 22 scenarios of fall captured by 8 cameras which include sequences of forward falls or backward falls while walking, falls when inappropriately sitting down, loss of balance etc. and 2 scenarios of normal daily activities such as walking in different directions, housekeeping, activities with characteristics similar to falls (sitting down/standing up, crouching down). The fall sequences in dataset are not trimmed action videos as they involve frames containing walking before fall, recovery phase and walking after fall. The temporal annotations of fall is also provided in the dataset which we use to create fall and non-fall sequences. 
The fall and non-fall video sequences from the first 17 scenarios along with 23rd scenario, are used as training set while the video sequences from 18th to 22nd along with 24th scenario, are used as test set.

In ~\cite{kwolek2014human}, the authors collected dataset containing 70 videos, comprising of 30 fall videos and 40 videos with activities of daily living. Fall and daily activities sequences were recorded with Microsoft Kinect cameras in form of RGB and depth data. Here we create the learning set containing 70 fall and 642 non-fall sequences with temporal strides. Fall sequences from first 24 fall videos and non-fall sequences from first 32 non-fall videos are used as training set and the rest are used as test set.

For pretraining our 3D CovNet, we create a learning set by randomly selecting 10 classes$^{*}$ from Kinetics-400 dataset ~\cite{carreira2017quo}. The actions involved in these 10 classes from Kinetics-400 are archery, belly dancing, cheerleading, dodgeball, high jump, playing cello, push up, swimming backstroke, tying tie and washing hair. This subset is composed of around 8K clips of YouTube videos. Each video includes only one actions. The training set, validation set and test set is divided as given in Kinetics-400 dataset.

\begin{table}[]
\centering
\caption{Accuracy on test split of Kinetics dataset with different deep learning architectures}
\begin{tabular}{|l|l|l|}
\hline
\textbf{Dataset} & \textbf{Method}                    & \textbf{Accuracy} \\ \hline
Kinetics-400     & I3D network  & 71.1\%            \\ 
                 & (ImageNet pre-trained) &               \\        \hline
Kinetics-10$^{*}$      & I3D network  & 92.3\%            \\ 
                 & (ImageNet pre-trained) &                \\\hline
Kinetics-10      & I3D network (scratch)              & 79.73\%           \\ \hline
\textbf{Kinetics-10}      & \textbf{3D ConvNet (scratch)}               & \textbf{78.98}\%           \\ \hline
\end{tabular}
\label{tab:kinetics_scratch}
\end{table}

\begin{table}[]
\centering
\caption{Accuracy on test split of Kinetics-10$^{*}$ with our 3D ConvNet architecture}
\label{tab:Sensing}

\begin{tabular}{|c|c|c|c|c|}
\hline
\multirow{2}{*}{\textbf{Sensing Matrix Type}} & \multicolumn{4}{c|}{\textbf{Compression Ratio}} \\ \cline{2-5} 
                                     & \textbf{4}             & \textbf{16}            & \textbf{32}     & \textbf{64}    \\ \hline
Random Gaussian Matrix               & 77.07         & 77.22         & \textbf{78.48}& \textbf{78.26} \\ \hline
Random Bernoulli Matrix              & 75.50         & 75.28         & 77.22         & 76.99 \\ \hline
Structural Measurement Matrix (SMM)~\cite{dinh2013measurement}  & \textbf{78.63}& 78.11         & 77.81         & 75.58 \\ \hline
Local Structural Measurement Matrix~\cite{gao2015block}  & 74.98         & 75.13         & 77.74         & 76.99 \\ \hline
Convolutional CS Measurement Matrix~\cite{shi2017deep}  & 77.96         & \textbf{78.78}& 76.62         & 75.23    \\ \hline
\end{tabular}%
\end{table}

Table \ref{tab:kinetics_scratch}, shows the accuracy performance on test split of Kinetics dataset with different deep learning architectures. Table ~\ref{tab:Sensing} shows the accuracy results over 10 classes of Kinetics dataset with random Gaussian, random Bernoulli, structural measurement matrix, local structural measurement matrix and Convolutional CS measurement matrix at different compression ratios. We train separately, from scratch, the 3D ConvNet for different compression ratios and different measurement matrices. The performance of 3D ConvNet is more or less similar for the reported measurement matrices. If we train I3D ~\cite{carreira2017quo} network from scratch over the given classes from Kinetics dataset, the performance comes out to be 79.73\% and the performance of our 3D ConvNet comes out to be 78.98\%. Since there is small difference in performance between I3D and our 3D ConvNet with compressive sensing, it is safe to say our 3D ConvNet is sufficient to learn actions for the reported action recognition dataset.


\begin{table}[]
\centering
\caption{Performance of various techniques over Multi-camera fall dataset and UR fall dataset}
\label{tab:fall}

\resizebox{\textwidth}{!}{\begin{tabular}{|c|c|c|c|c|}
\hline
\textbf{Method}                                                                                                 & \textbf{Compression ratio} & \textbf{Pre-trained on} & \textbf{Multi-camera } & \textbf{UR Fall dataset} \cite{kwolek2014human} \\ 
               &                             &  \textbf{Dataset}    & \textbf{fall dataset} & 
               \\\hline
Full Proscrustes distance  ~\cite{rougier2011robust}                                                                                     & 1 (No privacy)             & -                               & 96.20\%                            & -                        \\ \hline
3DCNN   ~\cite{lu2019deep}                                                                                                        & 1 (No privacy)             & Sports-1M  ~\cite{karpathy2014large}                     & 99.73\%                            & -                        \\ \hline
\begin{tabular}[c]{@{}c@{}}Visual Attention Guided\\ 3DCNN ~\cite{lu2019deep}\end{tabular}                                         & 1 (No privacy)             & Sports-1M  ~\cite{karpathy2014large}                     & 99.36\%                            & 99.27\%                  \\ \hline
\multirow{5}{*}{\begin{tabular}[c]{@{}c@{}}Proposed framework\\ (SMM+ 3DConv Inception\\ Network)\end{tabular}} & 1 (No privacy)             & Kinetics-10                     & 100\%                              & 100\%                    \\ \cline{2-5} 
                                                                                                                & 4                          & Kinetics-10                     & 100\%                              & 100\%                    \\ \cline{2-5} 
                                                                                                                & 16                         & Kinetics-10                     & 100\%                              & 100\%                    \\ \cline{2-5} 
                                                                                                                & 32                         & Kinetics-10                     & 100\%                              & 100\%                    \\ \cline{2-5} 
                                                                                                                & 64                         & Kinetics-10                     & 100\%                              & 100\%                    \\ \hline
    \end{tabular}}
\end{table}

In Table ~\ref{tab:fall}, we report the performance on fall detection dataset using pre-trained 3D ConvNet (over reported action recognition dataset) with structural measurement matrix at different compression ratios. Since fall detection is a binary classification problem, we report 100\% accuracy with pre-trained 3D ConvNet. We found that our 3D ConvNet architecture performs better than 3D CNN from \cite{lu2019deep} for fall detection.

\subsection{Implementation Details}
All action sequences (including fall and non-fall), were resized to 224x320 before compressed using measurement matrix. We train our model using ADAM optimizer with initial learning rate of $10^{-3}$ which is reduced by a factor of $10$ when validation loss doesn't decrease for $10$ consecutive epochs and training is terminated when validation loss doesn't decrease for 22 consecutive epochs. We implemented all the models in TensorFlow~\cite{tensorflow2015-whitepaper} and trained and evaluated them on nvidia-docker~\cite{nvidia_docker_2} for Tensorflow on NVIDIA DGX-1.

\section{Conclusion}
\label{sec:conclusion}
A compressive sensing based fall detection framework has been presented in the paper that also enables privacy preserving since it is a huge concern for patients being monitored through regular cameras. Our deep learning architecture performs similar to I3D network~\cite{carreira2017quo}, when trained from scratch, in accuracy for reported action recognition dataset, even with wide variety of compressive sensing measurement matrices. Experimental results on Multi-camera fall dataset and UR-Fall dataset were presented to show the effectiveness of the framework at different compression ratios. 

\section{Acknowledgment}
The NVIDIA DGX-1 for experiments was provided by CSIR-CEERI, Pilani, India

\bibliographystyle{splncs04}
\bibliography{refs}

\end{document}